\documentclass{article} 
\usepackage{iclr2026_conference,times}

\usepackage{amsmath,amsfonts,bm}

\def\eqref#1{equation~\ref{#1}}

\def\1{\bm{1}}

\DeclareMathAlphabet{\mathsfit}{\encodingdefault}{\sfdefault}{m}{sl}
\SetMathAlphabet{\mathsfit}{bold}{\encodingdefault}{\sfdefault}{bx}{n}

\usepackage{mwe}

\newlength{\tempdima}
\newcommand{\rowname}[1]
{\rotatebox{90}{\makebox[\tempdima][c]{\textbf{#1}}}}

\newcounter{subfigure}[figure]
\renewcommand{\thesubfigure}{\alph{subfigure}}
\newcommand{\mycaption}[1]
{\refstepcounter{subfigure}\textbf{(\thesubfigure) }{\ignorespaces #1}}

\usepackage{hyperref}
\usepackage{url}
\usepackage{graphicx}
\usepackage{booktabs} 

\usepackage{array,collcell}
\newcommand\AddLabel[1]{%
  \refstepcounter{equation}
  (\theequation)
  \label{#1}
}
\newcolumntype{M}{>{\hfil$\displaystyle}c<{$\hfil}} 
\newcolumntype{L}{>{\collectcell\AddLabel}r<{\endcollectcell}}

\title{Synaptic Activation and Dual Liquid Dynamics for Interpretable Bio-Inspired Models}

\author{Mónika Farsang \& Radu Grosu \\
Cyber-Physical Systems Research Unit\\
Vienna University of Technology (TU Wien)\\
Vienna, Austria \\
\texttt{\{monika.farsang,radu.grosu\}@tuwien.ac.at} \\
}

\iclrfinalcopy 
\begin{document}

\maketitle

\begin{abstract}
In this paper, we present a unified framework for various bio-inspired models to better understand their structural and functional differences. We show that liquid-capacitance-extended models lead to interpretable behavior even in dense, all-to-all recurrent neural network (RNN) policies. We further demonstrate that incorporating chemical synapses improves interpretability and that combining chemical synapses with synaptic activation yields the most accurate and interpretable RNN models. To assess the accuracy and interpretability of these RNN policies, we consider the challenging lane-keeping control task and evaluate performance across multiple metrics, including turn-weighted validation loss, neural activity during driving, absolute correlation between neural activity and road trajectory, saliency maps of the networks' attention, and the robustness of their saliency maps measured by the structural similarity index.
\end{abstract}

\section{Introduction}
Our work examines how concepts such as saturation (gating), input- and state-dependent capacitance mechanisms, and different activation types used in bio-inspired RNNs can be generalized to produce new model variants. We compare these to popular gated RNNs and other bio-inspired models. We show that liquid-capacitance mechanisms and chemical-synapse-based dynamics are essential for learning interpretable neural networks. Moreover, we find that biologically constrained, saturated (gated) RNNs are more interpretable than classic gated RNNs and that introducing synaptic activation further enhances interpretability.

%

%
%
Understanding and trusting the internal workings of gated RNNs remains challenging, especially in safety-critical applications, despite their success in modeling temporal dependencies. While these models share the goal of improving sequence learning, they vary substantially in architectural complexity. For instance, LSTMs \citep{10.1162/neco.1997.9.8.1735} incorporate both memory-cell and hidden-cell states with three gates (input, forget, and output), whereas GRUs \citep{cho2014learning} simplify this design by including only reset and update gates, maintaining performance with fewer parameters. MGUs \citep{zhou2016minimal} further reduce complexity by merging the reset and update gates into one.
Despite their practical advantages, these gated RNNs are difficult to relate directly to biological memory due to their abstract, mathematically oriented formulations. In contrast, Electrical-Equivalent Circuits (EECs) \citep{kandel2000principles, wicks1996dynamic} capture biological neuron behavior via electrical or chemical synapses, with each parameter having a clear biological interpretation. This attribute aligns well with the growing demand for interpretable AI systems.



Previous work explored EECs using either electrical~\citep{Funahashi1993ApproximationOD} or chemical synapses (also called Liquid Time-Constant networks (LTC)~\citep{Hasani2020LiquidTN, 2020NeuralCP}) in both sparse and fully connected architectures \citep{farsang2023learning}. It was shown that sparsity enhances interpretability and that chemical synapses are more robust than electrical ones. However, such studies primarily focused on Neural ODE formulations rather than gated (saturated) RNN extensions.

More recently, Liquid-Resistance Liquid-Capacitance networks (LRCs) \citep{farsang2024liquid, farsang2025parallelizationnonlinearstatespacemodels} have connected biological principles to efficient RNN design. These models build upon neuroscience-inspired mechanisms, integrating saturation (gating) and liquid capacitance into the chemical-synapse EECs to achieve both computational efficiency and biological plausibility. They operate with the combination of two liquid time-constants.


In this paper, we derive models with \textit{only} liquid-capacitance LCs from EECs with electrical synapses, analogous to how LRCs arise from EECs with chemical synapses. The electrical-synapse-based EECs are known to underlie continuous-time RNNs (CT-RNNs) \citep{Funahashi1993ApproximationOD, farsang2023learning}. The key difference between LRCs and LCs lies in their liquid-resistance: LRCs feature a state- and input-dependent forget gate (the liquid time constant), whereas LCs - based on electrical-synapse-based EECs - use a fixed term for the resistance.


We also demonstrate that synaptic activation in both LCs and LRCs further improves interpretability. In bio-inspired LRCs, activation functions are inherently associated with synapses, representing the probability that synaptic channels in postsynaptic neurons are open. This contrasts with artificial neural networks, where activations are applied per neuron. Nevertheless, if all outgoing synapses of a neuron share identical dynamic parameters, activations can be equivalently computed once per neuron, aligning with conventional ANN practice.


In summary, our contributions in this paper are as follows:
\begin{itemize}
%
\item We introduce new models with Liquid-Capacitance (LCs), which are formally derived from saturated EECs with electrical synapses. 

\item We show that LRCs
are more interpretable
than LCs, as they possess one more liquid gating mechanism, the liquid-resistance, which further regulates the amount of current state and update within the next state of these units.

\item We show that synaptic activation within LCs and LRCs further increases their interpretability, when compared to neural activation. 
%
\end{itemize}

The rest of the paper is organized as follows. 
%
%
In Section~\ref{sec:lcu}, we introduce LCs and review LRCs, with different activation types.
%
%
In Section~\ref{sec:experiments}, we conduct experiments to determine which features within the analyzed bio-inspired RNNs yield the most interpretable dynamics.  
Finally, we discuss our results in Section \ref{sec:discussion}.

\section{Liquid-Capacitance-based Models}\label{sec:lcu}


\begin{figure}[tb]
  \centering
  \includegraphics[width=\linewidth]{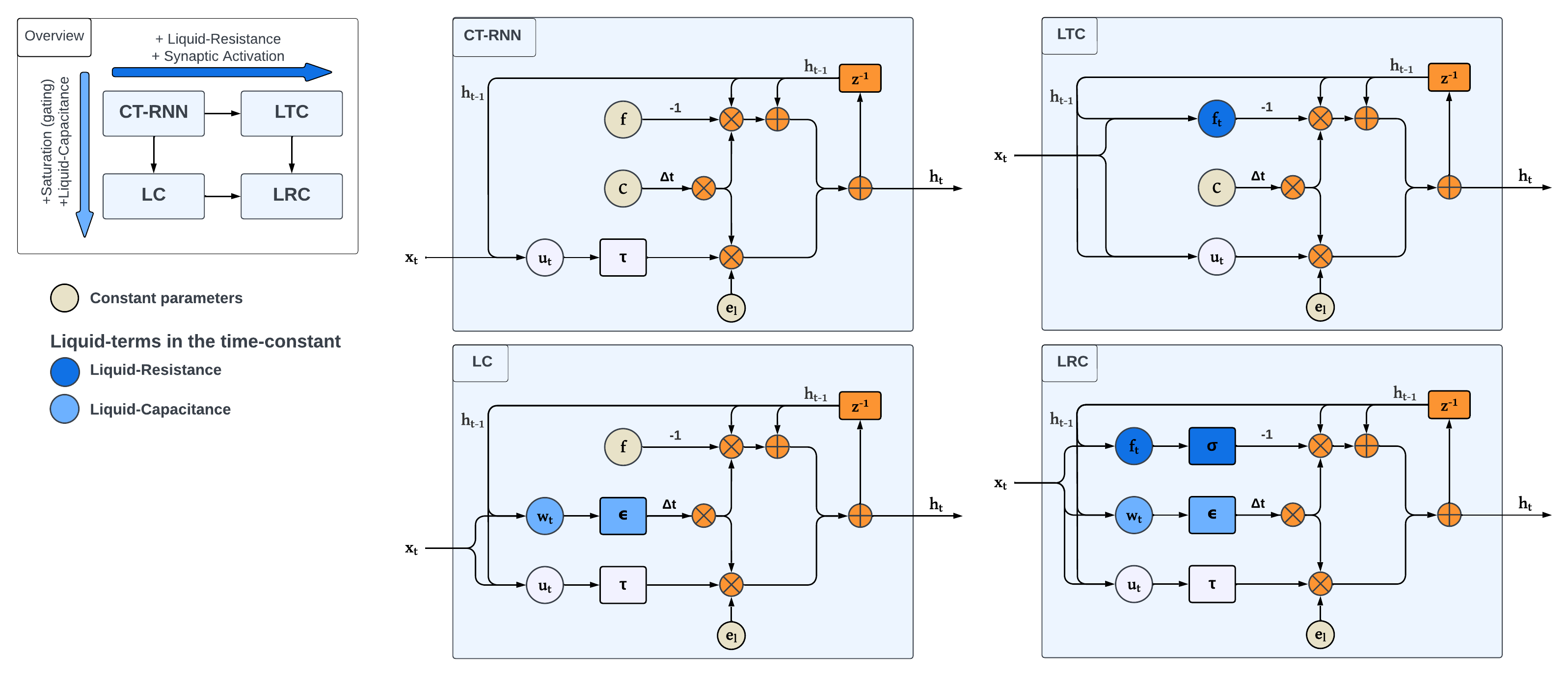}
  \caption{Structure of CT-RNNs and LTCs (top) and LC and LRC (bottom) cells, respectively. In addition to this, we further distinguish LC-NA and LC-SA, and LRC-NA and LRC-SA, respectively, to measure the effect of neural vs synaptic activation, which is incorporated in their $u$ and $f$ terms.
  }
  \label{fig:geu_gcu}
\end{figure}

%

We provide a brief overview here, about our unified framework, to apply the same concepts across different models types. Further details on the equations are summarized in Table~\ref{tab:f_u} and background on EECs in~\ref{sec:eecs}. All these models use Recurrence~\eqref{eq:LCU_discr}.
\subsection{Models with \textit{only} Liquid-Capacitance (LC)} 
We introduce models based on the simpler EEC architectures, that is, models with electrical synapses. Electrical synapses by default have neural activation, meaning that the outgoing synapses have the same dynamics, with no separate per pre-synaptic and post-synaptic neuron dynamics. CT-RNNs~\citep{Funahashi1993ApproximationOD} are based on this principle.

The contribution of these new models lies in the fact that they incorporate only liquid-capacitance dynamics (and no liquid-resistance) into their previously fixed time constants.

\paragraph{Liquid Capacitance with Neural Activation (LC-NA).} 
These LCs are biologically plausible, and therefore the default LCs. They employ the forget $f_i$ and update $u_i$ terms of~\eqref{eq:LC_NA_f_u}. LC-NA is an extended model of CT-RNNs~\citep{Funahashi1993ApproximationOD}. 

\paragraph{LC with synaptic activation (LC-SA).}
%
%
%
%
%
These LCs are not biologically plausible due to their synaptic activation. We introduce them in this paper for interpretability-ablation purposes, only. They employ the forget $f_i$ and update $u_i$ terms of~\eqref{eq:LC_SA_f_u}.              

\subsection{Models with Liquid-Resistance and Liquid-Capacitance (LRC)}

These models are based on chemical synapse EECs. They consider a \textit{double} liquid time-constant, with liquid-resistance and liquid-capacitance. Same as in~\citep{farsang2024liquid}.

\paragraph{LRC with neural activation (LRC-NA).}
%
%
%
%
These LRCs assume that all outgoing synapses of a neuron have the same dynamics. This might sometimes happen. They use the forget $f_i$ and update $u_i$ terms of~\eqref{eq:LRC_NA_f_u}.  

\paragraph{LRC with synaptic activation (LRC-SA).}
%
%
%
These LRCs are biologically plausible with their synaptic activation, and therefore, the default LRCs~\citep{farsang2024liquid}. They employ the forget $f_i$ and update $u_i$ terms of~\eqref{eq:LRC_SA_f_u}. This is the extended model of LTCs~\citep{Hasani2020LiquidTN}.

 
\begin{figure}[t]
  \centering
  \includegraphics[width=\linewidth]{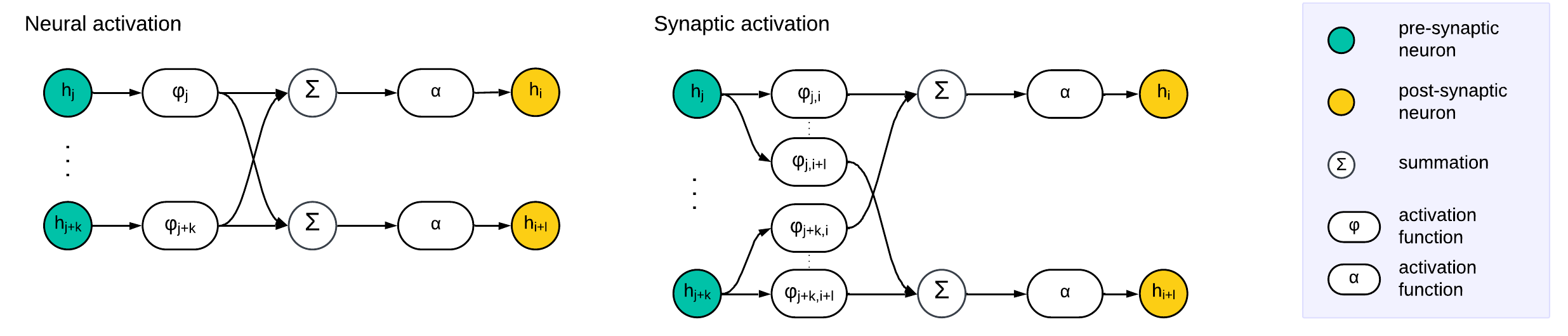}
  \caption{A neural network with either neural activation (NA) on the top, or synaptic activation (SA), on the bottom. While the NA model associates the activation $\varphi_j$ to the pre-synaptic neuron $h_j$ itself (by assuming that all outgoing synapses of neuron $h_j$ have the same dynamics), the SA model assumes different dynamics, by using separate activations $\{\varphi_{j,i}, ..., \varphi_{j,i+l}\}$ for each outgoing synapse from pre-synaptic $h_j$ to post-synaptic neurons $h_i$ to $h_{i+l}$.}
  \label{fig:activation_types}
  \vspace*{-2mm}
\end{figure}

\begin{figure}[tb]
  \centering
  \begin{minipage}[t]{0.35\linewidth}
    \centering
    \includegraphics[width=\linewidth]{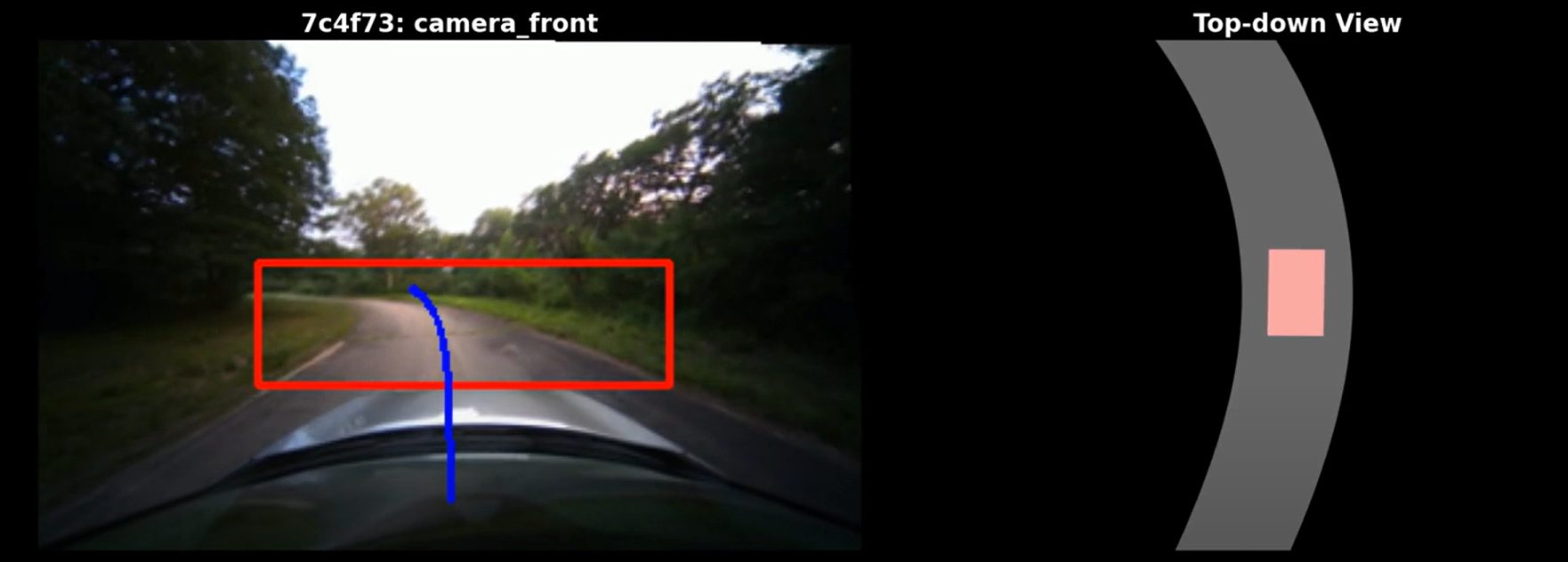}\\[0.5em]
    \includegraphics[width=\linewidth]{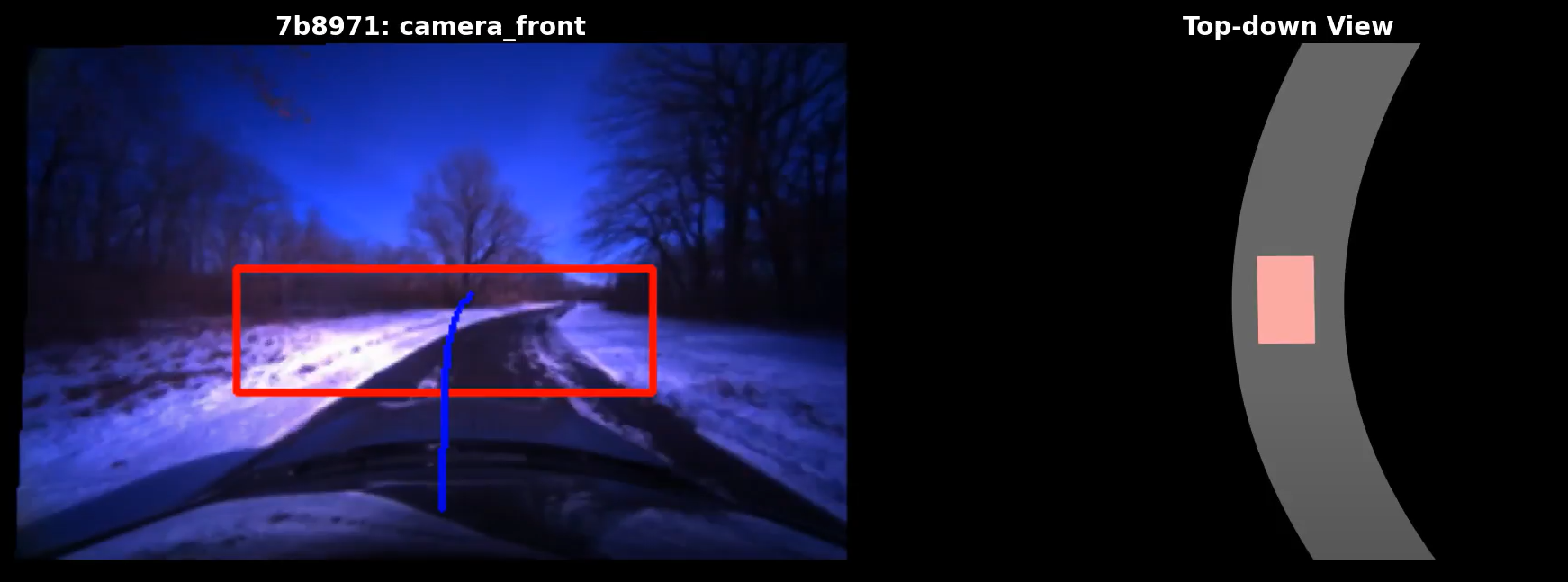}
  \end{minipage}
  \hfill
  \begin{minipage}[c]{0.55\linewidth}
    \centering
    \includegraphics[width=\linewidth]{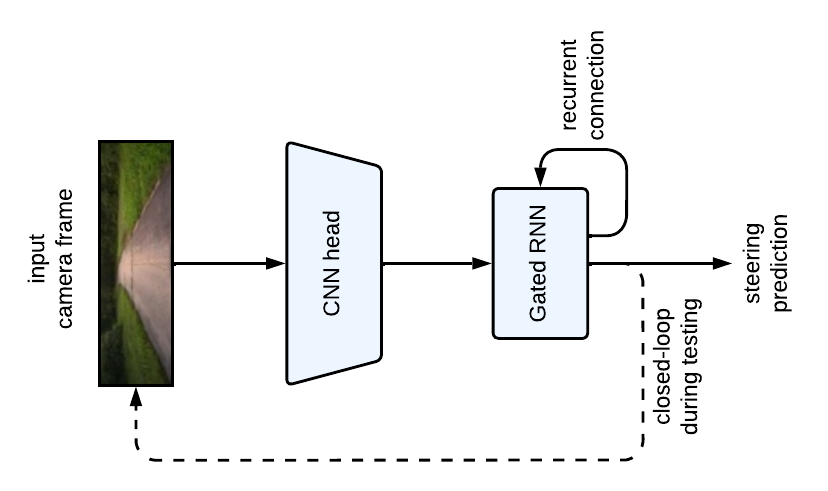}
  \end{minipage}

  \caption{
  Left: Simulation of autonomous driving in diverse seasons. The network's input is highlighted by the red rectangle, and its steering prediction is depicted by the blue spline. The car's position on the road is shown on the right side of the frames.
  Right: Starting from the front camera image, features are extracted using a CNN head and passed to an RNN policy to predict the steering angle. Training and validation are performed in an open-loop setting without the feedback connection (dashed line), which is only enabled during simulator testing.
  }
  \label{fig:vista_network}
\end{figure}

\section{Interpretability Evaluation}
\label{sec:experiments}
Determining the interpretability of a model is a tricky matter. On the one hand, the models should be accurate, that is, they should have a low validation loss, as otherwise they would not be interesting. On the other hand, the validation loss in itself, provides little information about the model's interpretability. Hence, one has to come up with metrics, specifically tailored to capture this property alone. The tasks solved by the models should also be relevant enough, and they should strike a good balance between simplicity and complexity. This way, one can still interpret the results, while hoping they generalize to more complex tasks, too.

\paragraph{Lane-Keeping task.} 
We decided to conduct our experiments on the autonomous end-to-end Lane-Keeping task from \citet{lechner2022all}. In this task, the networks receive sequential, front-camera input from a car, and have to generate sequential steering commands to keep the car on the track. We utilize imitation learning to train the agents, with expert demonstrations provided by a human driver. After training the networks, we evaluate them in the data-driven VISTA simulator \citep{Amini2020LearningRC, amini2022vista}, to assess their performance and interpretability in a closed-loop setting. The simulation environment is illustrated in Figure \ref{fig:vista_network}, where we showcase two different seasons, summer and winter.

The agent receives input from the red rectangle portion of the image, sized to $48{\times}160$ pixels. Its predictions are represented by the blue line, while on the right-hand side, the car's position on the road is visualized from a bird's-eye-view perspective. 
As shown in Figure~\ref{fig:vista_network}, the pixel input is fed into a convolutional neural network (CNN), which outputs the feature maps of the image. These features serve as input to the RNN, responsible for predicting the steering angle. The examined recurrent networks (RNNs), including the traditionally-used gated RNNs, and liquid capacitance units, are incorporated into a fully-connected RNN architecture of 19 neurons.

\begin{table}[t]
    \centering
     \caption{Absolute correlation values between the road's trajectory and the activities of all neurons in the networks. LRC-NA and LRC-SA demonstrate more than 1.5x better correlations compared to the other models. Results are averaged over 3 runs. }
    \begin{tabular}{lcc}
    \toprule
         & Summer & Winter\\
     \midrule
        LSTM & $0.450 \pm 0.282$ & $0.434 \pm 0.285$\\
        GRU & $0.395 \pm 0.269$ & $0.368 \pm 0.274$\\
        MGU & $0.383 \pm 0.253$ & $0.411 \pm 0.269$\\
        CT-RNN & $0.390 \pm 0.267$ & $0.315 \pm 0.243$ \\
        LTC & $\mathbf{0.666 \pm 0.296}$ & $0.662 \pm 0.274$ \\
        LC-NA   & $0.477 \pm 0.243$ & $0.428 \pm 0.254$\\
        LC-SA   & $0.462 \pm 0.273$ & $0.533 \pm 0.246$\\
        LRC-NA   & $\mathbf{0.652 \pm 0.293}$ & $\mathbf{0.693 \pm 0.292}$\\
        LRC-SA  & $\mathbf{0.645 \pm 0.321}$ & $\mathbf{0.766 \pm 0.243}$\\
     \bottomrule
    \end{tabular}
    \label{tab:correlations}
\end{table}

\begin{figure*}[t]
  \centering
  \includegraphics[width=0.49\linewidth]{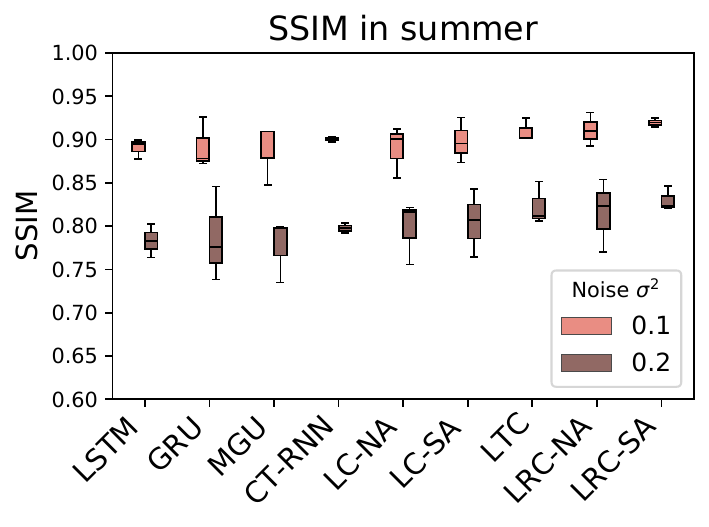}
  \includegraphics[width=0.49\linewidth]{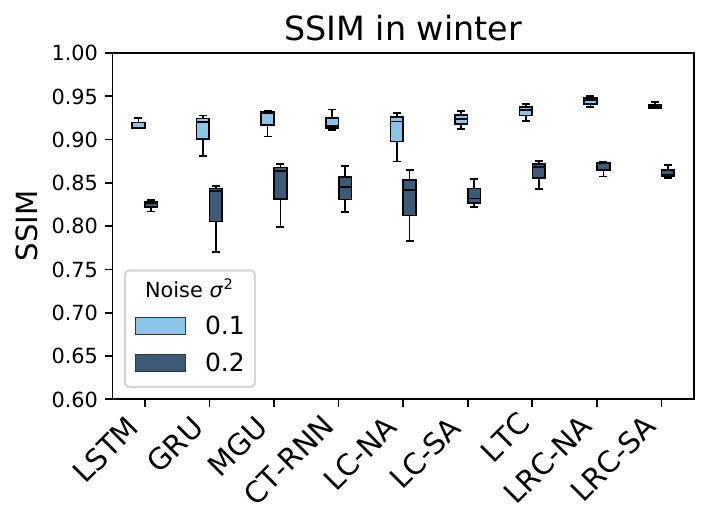} 
  \caption{Robustness of the attention, measured by the Structural Similarity Index (SSIM) of the models, in summer (left) and in winter (right). The LRC-based models maintain the most similar focus of their attention in the presence of noise, indicated by the boxplots closer to 1. Lighter color refers to additional Gaussian noise of zero-mean and $\sigma^2=0.1$ variance, and darker to $\sigma^2=0.2$ variance.}
  \label{fig:ssim}
\end{figure*}

\paragraph{Neural activity.}
As one metric of interpretability, 
we assess how the neural activity of the neurons within the RNN Lane-Keeping policy changes during deployment in the closed-loop simulation. Specifically, we are interested in identifying neurons that exhibit an activity which matches the geometry of the trajectory. We conduct tests for a 1 km long drive, in both summer and winter seasons. In Figure~\ref{fig:neural_activity}, we illustrate the activity of state $h_1$ and $h_2$, corresponding to neurons 1 and 2 in the RNN policy, respectively. Our findings reveal that all RNNs exhibit a degree of interpretability, with their neural activity matching more or less the changes in trajectory geometry. However,  the LSTM, GRU, MGU, CT-RNN and LC neurons, respectively, still have a somewhat scattered activity. In contrast, the LTC and LRC neurons demonstrate a much smoother, continuously changing activity, that aligns very well with the road's trajectory. This is most likely due to their liquid-resistance. 

As shown in Table~\ref{tab:correlations}, we also compute the absolute-value cross-correlation, between the final prediction sequence of the policy (which aligns with the road's trajectory), and the sequence of activities of individual neurons. By considering the absolute value of the cross-correlations, we ensure that equal importance is given to both positive (that is excitatory) and negative (that is inhibitory) behaviors. This method yields values within the range of $[0, 1]$, where values close to zero indicate little to no correlation, and values close to one mean a very high correlation. The results presented in Table~\ref{tab:correlations} reinforce the observations from Figure \ref{fig:neural_activity}, indicating that LRC-based models exhibit higher absolute correlation values with the road's trajectory compared to other models.

\paragraph{Network attention}
Understanding the attention of the neural network during decision-making, is a crucial element in increasing the trust in the network. By using the VisualBackprop \citep{Bojarski2016VisualBackPropVC} method, we visualize which pixels of the input image have the most impact at the given timestep. Results and their discussion is in Appendix~\ref{app:saliency}.

By measuring the change of attention, we also compute the Structural Similarity Index (SSIM) \citep{SSIM} of the images pairwise, between the noise-free and the noisy saliency map for each model, respectively. This technique assesses the similarity between two images, with a focus on image degradation. An SSIM value of one denotes full similarity, whereas one of zero, denotes zero similarity. In our particular case, where we are comparing attention-map images, values close to one mean that the extra noise leads to less distortion in the attention map, which is a very desirable aim in the development of robust controllers. The SSIM value for $1600{-}1600$ pairs of images, used in the attention-map comparisons between noise-free and noisy, summer and winter images, respectively, are presented in Figure~\ref{fig:ssim}. All results in this table are averaged over three runs for each model. As one can see in the figure, LCs exhibit a performance comparable to the one of traditionally-used gated recurrent units. However, LRCs demonstrate a much more robust attention in the presence of Gaussian noise.

\paragraph{Synaptic versus Neural activation.}
Our analysis of interpretability for dense RNNs so far, has shown that LRC-NAs and LRC-SAs provide the best performance in terms of validation loss and robustness to noise. Moreover, their neural activity on the road is also the most interpretable in terms of their driving decisions. This is most likely due to their same gating mechanism, where the forget gate represents the liquid time constant of their associated EECs. 

\begin{figure}[h]
  \centering
  \begin{minipage}{0.45\linewidth}
    \centering
    LRC-NA\\
    \includegraphics[width=\linewidth]{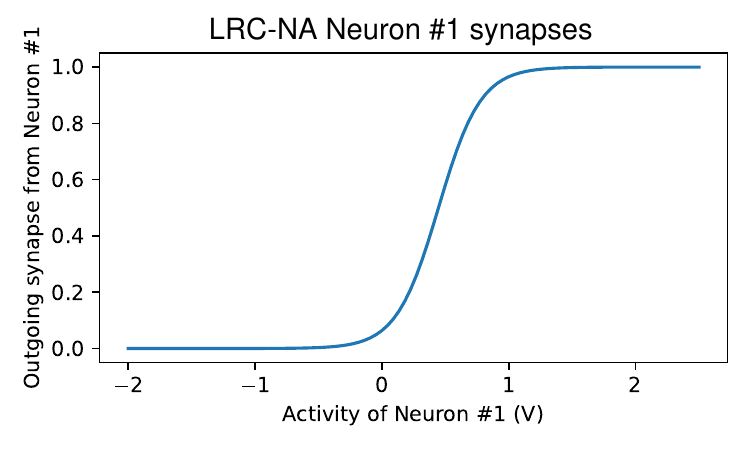}
  \end{minipage}
  \begin{minipage}{0.45\linewidth}
    \centering
    LRC-SA\\
    \includegraphics[width=\linewidth]{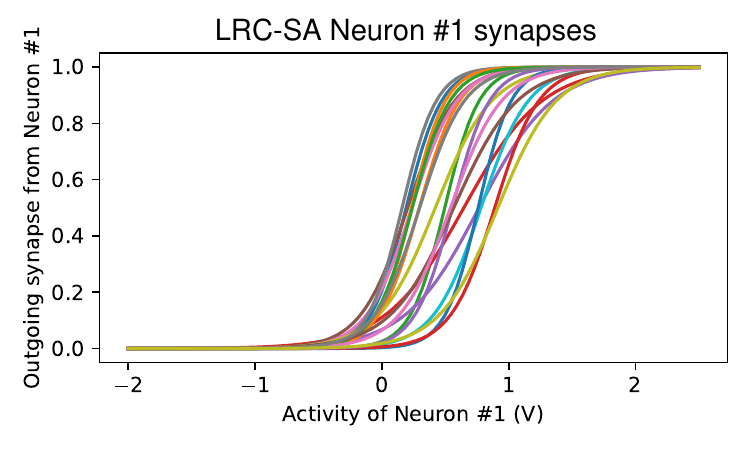}
  \end{minipage}
  \caption{Sigmoidal-activation functions learned for Neuron~1 of the Lane-Keeping policy of LRC-NA and LRC-SA. As one can see, Neuron~1 in LRC-SA takes advantage of its additional flexibility to learn a different dynamics for each of its outgoing synapses.}
  \label{fig:gcu_synapses}
\end{figure}

In this section, we 
further investigate whether their neural versus synaptic activation mechanism, also makes a difference in terms of interpretability. As illustrated in Figure~\ref{fig:gcu_synapses}, LRC-SA has the additional flexibility of learning a different dynamics (different parameters) for each outgoing synapse of a neuron. The question is whether the LRC-SA model would not automatically discover during training, that it was sufficient to learn just one dynamics per neuron, and thus essentially converge to the neural activation of a LRC-NA model?
%
%
We found out experimentally in the Lane-Keeping task, that LRC-SAs actually take advantage of the additional flexibility, as shown in Figure~\ref{fig:gcu_synapses}, by learning different activation parameters for different outgoing synapses. 

The added value to interpretability of this different LRC-SA synaptic-dynamics behavior, is best appreciated in Table~\ref{tab:correlations} and in Figure~\ref{fig:ssim}. As one can see in Table~\ref{tab:correlations}, LRC-SA and LRC-NA have essentially the same absolute correlation value in summer, but LRC-SA is considerably superior in winter. Similarly, as one can see in Figure~\ref{fig:ssim}, LRC-SA has the best SSIM value, with the smallest variance, in summer, and a similar performance with the LRC-NA in winter.

\section{Discussion}\label{sec:discussion}
Our results show that dual liquid dynamics (both liquid-resistance and liquid-capacitance) lead to interpretability in dense all-to-all RNN policies, with LRCs having the best validation losses, and also being the most interpretable among all RNNs. Moreover, by combining chemical-synapse gating with synaptic activation in LRCs, leads to learned models which have an even more explainable dynamics, and an even more focused attention, which is also more robust when subjected to additional perturbations. 
%

This work contributes to the foundations for concept learning with interpretable, bio-inspired models. We used a supervised approach to learn meaningful concepts directly from raw camera-stream inputs and evaluated the quality of these learned concepts using dedicated metrics quantitatively, with a strong focus on how well they support the downstream lane-keeping task.

For future work, we would like to perform additional experiments with these models, in diverse, and similarly challenging settings, where network interpretability is crucial, as is the case in safety-critical applications. Moreover, inspired by neuroscience and building on the findings in this paper, we would like to further explore the nature of these neural models, and build interpretable networks in a modular fashion.


\subsubsection*{Acknowledgments}
This project has received funding from the European Union’s Horizon 2020 research and innovation programme under the Marie Skłodowska-Curie grant agreement No 101034277.

\bibliography{refs}
\bibliographystyle{iclr2026_conference}

\newpage
\appendix
\section{Background on Electrical Equivalent Circuits}\label{sec:eecs}

The Electrical-Equivalent-Circuits model (EECs) used in neuroscience, capture the dynamic behavior of biological neurons with both chemical and electrical synapses, in form of neural ordinary differential equations (Neural ODEs), as described in \citet{kandel2000principles,wicks1996dynamic}. 

\paragraph{Membrane potentials (MP).} Neurons are the basic building blocks of the nervous system and are responsible for transmitting and processing signals. 
In EECs, a neuron-$i$'s membrane is regarded as a capacitor whose potential $h_i$ is determined by the difference between intra- and extra-cellular concentrations of ions, respectively. The membrane potential $h_i$ thus dynamically evolves as follows:
\begin{equation}
    C_i\,\dot{h}_i = i_{li} + i_{si},\quad i_{li} = g_{li}(e_{li} - h_i)
\label{eq:eec}
\end{equation}
%
%
where $C$ is the membrane's capacitance, and $i_{li}$ and $i_{si}$ are the passive-leaking and dynamic-synaptic currents, traversing the membrane, respectively. We will detail the capacitance term in the next section~\ref{sec:capacitance}. The leaking current is determined by the Ohm's equation, where $e_{li}$ is the resting (or reversal) potential of the neuron, and $g_{li}$ is its leaking conductance. They are considered to be constant.

\paragraph{Electrical synapses (ES).} 
The prevalent model of ESs is an Ohmic current $i_{s,ji}\,{=}g_{ji}(y_j\,{-}\,h_i)$, where $y_j$ is either the MP $h_j$ of a presynaptic neuron, or the MP $x_j$ of an input. Constant $g_{ji}$ is the conductance of the synapse. Hence, in electrical synapses, $h_i$ faithfully follows $y_j$. Let $y\,{=}[h,x]$, $|h|\,{=}\,m$ and $|x|\,{=}\,n$. As a network, one thus obtains:
\begin{equation}
    \dot{h}_i =g_{li}(e_{li}-h_i) + \sum_{j=1}^{m+n} g_{ji} (y_j-h_i)
\label{eq:ES1}
\end{equation}
\vspace*{-8mm}\paragraph{Chemical Synapses (CS).} The CS model is a current $i_{s,ji}\,{=}g_{ji}\sigma(a_{ji}y_j+b_{ji})(e_{ji}\,{-}\,h_i)$, passing through the neuron-$i$'s channels, when these open, as neurotransmitters released by presynaptic neurons-$j$ start binding to the receptors of the channels. Here, $g_{ji}$ is the maximum conductance of the channels, $\sigma(a_{ji}y_j+b_{ji})$ the probability of the channels to be open, with parameters $a_{ji}$ and $b_{ji}$, and $e_{ji}$ is the reversal potential of the channels. 
A network is thus:
\begin{equation}
    \dot{h}_i =g_{li}(e_{li}-h_i) + \sum_{j=1}^{m+n} g_{ji}\sigma(a_{ji}y_j+b_{ji})(e_{ji}-h_i)
\label{eq:CS1}
\end{equation}
\paragraph{Saturated normal form (SEEC).}
Rearranging the terms in the ESs and CSs, and saturating conductances, one obtains the following normal form of the saturated EECs, in \eqref{eq:ES2} and \eqref{eq:sltc}, respectively. Since $f_i$ is a conductance, that cannot grow unbounded in biological neurons, it is saturated with a sigmoid function $\sigma$. Similarly, the conductance $u_i$ is saturated with a hyperbolic tangent $\tau$, because $k_{ji}\,{=}\,g_{ji}{/}e_{li}$ takes the sign of $e_{ji}$.
\begin{equation}
\begin{array}{c}
    \dot{h}_i =-\sigma(f_i)\,h_i + \tau(u_i)\,e_{li}\\[2mm]
    f_i = g_{li}+\sum_{j=1}^{m+n} g_{ji}\\[2mm]
    u_i = g_{li}+\sum_{j=1}^{m+n} k_{ji}y_j
\label{eq:ES2}
\end{array}
\end{equation}
Since $\sigma(f_i)$ is a constant, the EEC model for ESs is identical to CT-RNNs, modulo $e_{li}$~\citep{Funahashi1993ApproximationOD}, also the same as what we will use in \eqref{eq:LC_NA_f_u}.
\begin{equation}
\label{eq:sltc}
\begin{array}{c}
    \dot{h}_i =-\sigma(f_i)\,h_i + \tau(u_i)\,e_{li}\\[2mm]
    f_i = g_{li}+\sum_{j=1}^{m+n} g_{ji}\sigma(a_{ji}y_j+b_{ji})\\[2mm]
    u_i = g_{li}+\sum_{j=1}^{m+n} k_{ji}\sigma(a_{ji}y_j+b_{ji})
\end{array}
\end{equation}
In both EECs, the term multiplying the state $h_i$ is called the time constant of the EEC. For CSs, this term is liquid, that is, a function of input and state. Thus, the EECs for CSs are also called liquid time constant networks (LTCs) in~\citet{worm_inspired,Hasani2020LiquidTN,2020NeuralCP}.

\paragraph{Neural or synaptic activation.}
By default, CSs have an activation per synapse, and ESs have an activation per neuron. However, if all outgoing synapses of a neuron have the same dynamics, one can compute their activation, once per neuron, too. Thus, their conductances before saturation take the form of \eqref{eq:LRC_NA_f_u}.
%

To elucidate the role of saturation and synaptic/neural activation in interpretability, we also introduce a synaptic-activation form for ES. The associated equations are \eqref{eq:LC_SA_f_u}.  
%
%
\paragraph{Membrane Capacitance}~\label{sec:capacitance}
%


In~\eqref{eq:elastance_normal_distr} below, we show the symmetric-elastance definition of~\citep{farsang2024liquid}, which we use in this paper to model input- and state-dependent \textit{liquid} capacitance.
\begin{equation}
\label{eq:elastance_normal_distr}
\begin{array}{c}
\epsilon_{i,t} = \sigma (w_{i,t} + k_i) - \sigma (w_{i,t} - k_i)\\[2mm]
w_{i,t} = \sum_{j=1}^{m+n} o_{ji} y_{j,t} + p_{j}
\end{array}
\end{equation}

All the saturated normalized EEC equations discussed in this section, are summarized in Table~\ref{tab:f_u}.

\begin{table}[h]
    \centering
    \caption{Ordinary differential and difference equation of all models: LC-NA, LC-SA, LRC-NA, and LRC-SA. While they share the same structure, the models employ different functions in the sigmoid $\sigma(f)$ and hyperbolic tangent $\tau(u)$ term. Below: Forget $f$ and update $u$ equations characterizing the analysed models. LCs are electrical-synapse-based. They are given in the first two rows, for neural and synaptic activation, respectively. 
    LRCs are chemical-synapse-based.
    The last two rows describe their equations.  }
         \begin{tabular}{cL}
     \toprule
          Base Equations &  \multicolumn{1}{l}{}\\
     \midrule
         $\dot{h}_i =-\sigma(f_i)\epsilon_i\,h_i + \tau(u_i)\epsilon_i\,e_{li}$ & eq:LCU_ode \\
         \midrule
      $h_{i,t} = (1-\sigma(f_{i,t})\,\epsilon_{i,t}\Delta t)\,h_{i,t-1} + \tau(u_{i,t})\,\epsilon_{i,t}\,e_{li}\Delta t$ & eq:LCU_discr\\
      \bottomrule
     \end{tabular}
    \begin{tabular}{lMML}
    \toprule
         & f_i & u_i &  \multicolumn{1}{l}{}\\
    \midrule
        LC-NA & f_i = g_{li}+\sum_{j=1}^m g_{ji} & u_i = g_{li}+\sum_{j=1}^m k_{ji}y_j & eq:LC_NA_f_u \\
        LC-SA &  f_i = 
        g_{li}+\sum_{j=1}^m g_{ji} & 
        u_i = g_{li}+\sum_{j=1}^{m+n} k_{ji}\sigma(a_{ji}y_j+b_{ji}) & eq:LC_SA_f_u \\[2mm]
        \hline\\[-2mm]
        LRC-NA & f_i = g_{li}+\sum_{j=1}^{m+n} g_{ji}\sigma(a_j y_j+b_j) & u_i = g_{li}+\sum_{j=1}^{m+n} k_{ji}\sigma(a_j y_j+b_j) & eq:LRC_NA_f_u\\
        LRC-SA & f_i = g_{li}+\sum_{j=1}^{m+n} g_{ji}\sigma(a_{ji}y_j+b_{ji}) & u_i = g_{li}+\sum_{j=1}^{m+n} k_{ji}\sigma(a_{ji}y_j+b_{ji}) & eq:LRC_SA_f_u \\
     \bottomrule
    \end{tabular}
    \label{tab:f_u}
\end{table}

 
\newpage
\section{Training Details}

\paragraph{Network training.} We split the dataset into training, validation, and test sets. We use both seasons for each part of the model-training pipeline. During training, we use an open-loop offline imitation-learning approach, where the model receives a sequence of input images, both from the summer and the winter seasons. At each time point $t$, the RNN policy has to predict the steering angle $y_t$, which does not influence the subsequent input image. The training lasts for 100 epochs, where we choose the best validation loss, without early stopping \citep{Power2022GrokkingGB}. A sequence of 32 images is used in the batch size of 32 during training, where the AdamW~\citep{Loshchilov2017FixingWD} optimizer is applied with the learning rate of $5 \cdot 10^{-4}$, and weight decay of $10^{-6}$. During testing, the data-driven VISTA simulator generates new viewpoints based on past steering commands, resulting in closed-loop testing. At this stage, we save the online activities of neurons for further analysis. 

\paragraph{Validation loss.}
 For calculating the validation loss, we use the mean-squared error between the predicted and the ground-truth steering angle. We observed that the models learn different turning behavior, which can be highlighted by calculating a second metric, called weighted validation loss. Here, the assigned weights are proportional to the steepness of the turn, making stronger turns more influential. Table~\ref{tab:val_losses} reports our results. As illustrated in this table, the LRC-based models converge to the lowest validation loss, in half of the time, compared to the other models. They are also the strongest ones in terms of validation and weighted validation loss. Interestingly, LSTMs have the same weighted validation loss, even though their validation loss is higher.
\begin{table*}[h]
    \centering
    \caption{Validation and weighted validation losses of models. Experiments are repeated 3 times.}
    \begin{tabular}{lccc}
    \toprule
         & Best epoch & Validation Loss & Weighted Validation Loss\\
     \midrule
        LSTM & $60 \pm 17$ & $0.145 \pm 0.002$ & $\mathbf{0.011 \pm 0.001}$\\
        GRU & $52 \pm 13$ & $0.146 \pm 0.003$ & $0.012 \pm 0.001$\\
        MGU & $70 \pm 14$ & $0.145 \pm 0.007$ & $0.013 \pm 0.001$\\
        CT-RNN & $67 \pm 30$ & $0.218 \pm 0.019$ & $0.018 \pm 0.002$\\
        LTC & $59 \pm 12$ & $0.181 \pm 0.015$ & $\mathbf{0.010 \pm 0.0004}$\\
        LC-NA & $52 \pm 14$  & $0.173 \pm 0.011$ & $0.015 \pm 0.0004$\\
        LC-SA & $46 \pm 4$  & $0.163 \pm 0.004$ & $0.015 \pm 0.0004$\\
        LRC-NA & $34 \pm 7$  & $\mathbf{0.139 \pm 0.008}$ & $\mathbf{0.011 \pm 0.0004}$\\
        LRC-SA & $22 \pm 3$ & $\mathbf{0.142 \pm 0.004}$ & $\mathbf{0.011 \pm 0.001}$\\
     \bottomrule
    \end{tabular}
    \label{tab:val_losses}
\end{table*}

\paragraph{Trade offs.} We analyzed the trade-off with respect to network interpretability between architectures using the same number of neurons (and varying trainable parameters) and those using approximately the same number of trainable parameters (with a varying number of neurons). Traditional gated RNNs and LCs are more efficient in terms of parameters. Thus, to arrive at the same number of trainable parameters as the ones in the LRC-SA model, one should incorporate more neurons in the networks, to match the computational capacity. However, we surprisingly found that, when it comes to interpretability, this actually diminishes the interpretability of those networks. The reason is that a relatively small number of neurons forces the network to learn more meaningful features per neuron, compared to what happens when more neurons are incorporated. This realization led us to the decision to limit all these networks to 19 neurons, only, aiming to facilitate interpretable representations of this task, even though it did not precisely match the number of trainable parameters per model type. 

\newpage
\section{Additional Interpretability Results}
\subsection{Saliency Maps}~\label{app:saliency}
Even though the same structure of CNN heads is used in the training pipeline, the different recurrent parts of the decision-making, influence the learned features in the convolutional part through backpropagation, too.

\begin{figure}[h]
\settoheight{\tempdima}{\includegraphics[width=.05\linewidth]{example-image-a}}%
\centering\begin{tabular}{@{ }c@{ }c@{ }c@{ }c@{ }c@{ }c@{ }c@{ }c@{ }c@{ }c@{ }c@{ }}
 & & LSTM & GRU &  MGU & CT-RNN & LTC & LC-NA & LC-SA & LRC-NA & LRC-SA \\
\rowname{0.0}&
\includegraphics[width=.090\linewidth]{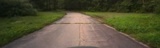}&
\includegraphics[width=.090\linewidth, decodearray={0.25 1 0.25 1 0.25 1}]{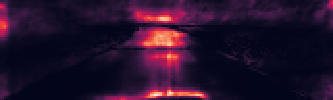}&
\includegraphics[width=.090\linewidth, decodearray={0.25 1 0.25 1 0.25 1}]{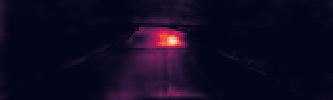}&
\includegraphics[width=.090\linewidth, decodearray={0.25 1 0.25 1 0.25 1}]{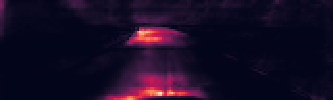}&
\includegraphics[width=.090\linewidth, decodearray={0.25 1 0.25 1 0.25 1}]{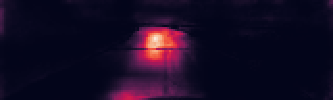}&
\includegraphics[width=.090\linewidth, decodearray={0.25 1 0.25 1 0.25 1}]{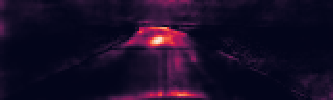}&
\includegraphics[width=.090\linewidth, decodearray={0.25 1 0.25 1 0.25 1}]{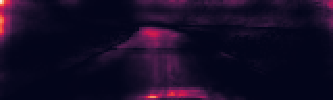}&
\includegraphics[width=.090\linewidth, decodearray={0.25 1 0.25 1 0.25 1}]{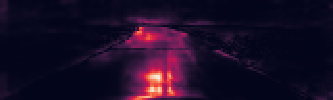}&
\includegraphics[width=.090\linewidth, decodearray={0.25 1 0.25 1 0.25 1}]{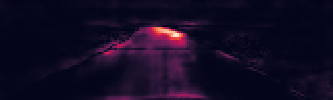}&
\includegraphics[width=.090\linewidth, decodearray={0.25 1 0.25 1 0.25 1}]{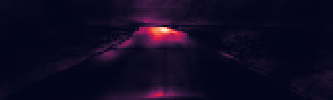}\\
\rowname{0.1}&
\includegraphics[width=.090\linewidth]{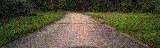}&
\includegraphics[width=.090\linewidth, decodearray={0.25 1 0.25 1 0.25 1}]{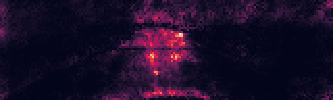}&
\includegraphics[width=.090\linewidth, decodearray={0.25 1 0.25 1 0.25 1}]{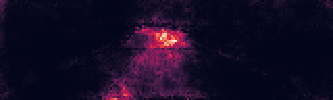}&
\includegraphics[width=.090\linewidth, decodearray={0.25 1 0.25 1 0.25 1}]{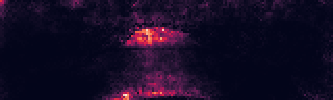}&
\includegraphics[width=.090\linewidth, decodearray={0.25 1 0.25 1 0.25 1}]{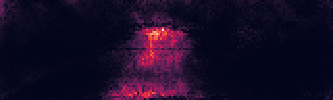}&
\includegraphics[width=.090\linewidth, decodearray={0.25 1 0.25 1 0.25 1}]{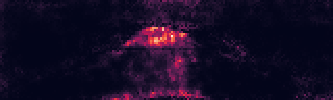}&
\includegraphics[width=.090\linewidth, decodearray={0.25 1 0.25 1 0.25 1}]{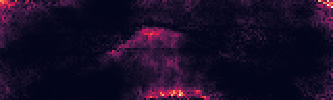}&
\includegraphics[width=.090\linewidth, decodearray={0.25 1 0.25 1 0.25 1}]{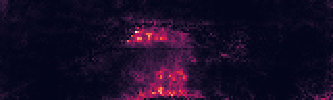}&
\includegraphics[width=.090\linewidth, decodearray={0.25 1 0.25 1 0.25 1}]{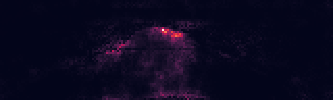}&
\includegraphics[width=.090\linewidth, decodearray={0.25 1 0.25 1 0.25 1}]{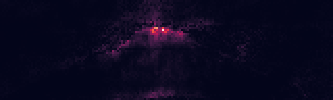}\\
\rowname{0.2}&
\includegraphics[width=.090\linewidth]{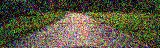}&
\includegraphics[width=.090\linewidth, decodearray={0.25 1 0.25 1 0.25 1}]{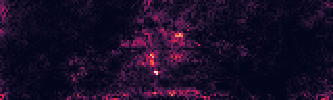}&
\includegraphics[width=.090\linewidth, decodearray={0.25 1 0.25 1 0.25 1}]{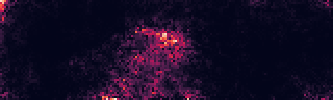}&
\includegraphics[width=.090\linewidth, decodearray={0.25 1 0.25 1 0.25 1}]{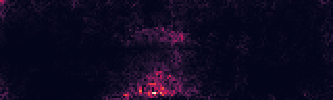}&
\includegraphics[width=.090\linewidth, decodearray={0.25 1 0.25 1 0.25 1}]{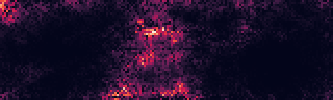}&
\includegraphics[width=.090\linewidth, decodearray={0.25 1 0.25 1 0.25 1}]{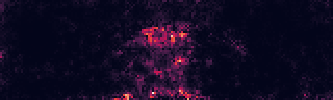}&
\includegraphics[width=.090\linewidth, decodearray={0.25 1 0.25 1 0.25 1}]{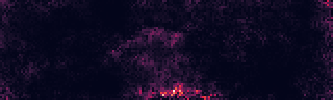}&
\includegraphics[width=.090\linewidth, decodearray={0.25 1 0.25 1 0.25 1}]{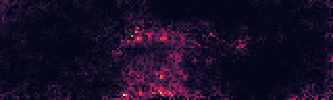}&
\includegraphics[width=.090\linewidth, decodearray={0.25 1 0.25 1 0.25 1}]{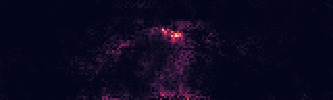}&
\includegraphics[width=.090\linewidth, decodearray={0.25 1 0.25 1 0.25 1}]{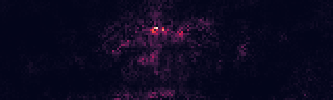}\\
\end{tabular}
\caption{Attention of the networks in summer. Column~1 shows the input of the network: Row 1, without additional noise, Rows 2-3, with Gaussian noise of variance $\sigma^2=0.1$ and $\sigma^2=0.2$, respectively. Remaining columns correspond to the networks considered, showing their attention to the same input image. One can observe how much the focused areas, get distorted in the presence of noise.}%
\label{fig:saliency_summer}
\end{figure}

\begin{figure}[h]
\settoheight{\tempdima}{\includegraphics[width=.05\linewidth]{example-image-a}}%
\centering\begin{tabular}{@{ }c@{ }c@{ }c@{ }c@{ }c@{ }c@{ }c@{ }c@{ }c@{ }c@{ }c@{ }}
 & & LSTM & GRU &  MGU & CT-RNN & LTC & LC-NA & LC-SA & LRC-NA & LRC-SA \\
\rowname{0.0}&
\includegraphics[width=.090\linewidth]{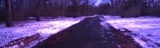}&
\includegraphics[width=.090\linewidth, decodearray={0.25 1 0.25 1 0.25 1}]{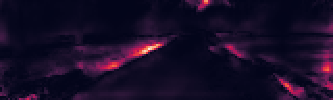}&
\includegraphics[width=.090\linewidth, decodearray={0.25 1 0.25 1 0.25 1}]{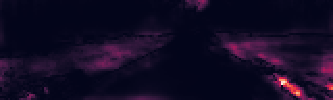}&
\includegraphics[width=.090\linewidth, decodearray={0.25 1 0.25 1 0.25 1}]{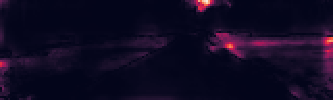}&
\includegraphics[width=.090\linewidth, decodearray={0.25 1 0.25 1 0.25 1}]{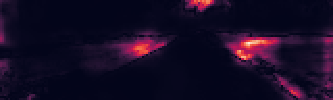}&
\includegraphics[width=.090\linewidth, decodearray={0.25 1 0.25 1 0.25 1}]{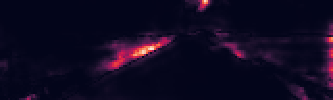}&
\includegraphics[width=.090\linewidth, decodearray={0.25 1 0.25 1 0.25 1}]{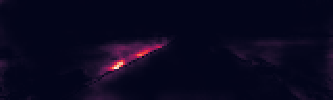}&
\includegraphics[width=.090\linewidth, decodearray={0.25 1 0.25 1 0.25 1}]{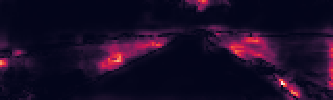}&
\includegraphics[width=.090\linewidth, decodearray={0.25 1 0.25 1 0.25 1}]{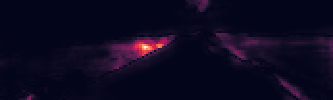}&
\includegraphics[width=.090\linewidth, decodearray={0.25 1 0.25 1 0.25 1}]{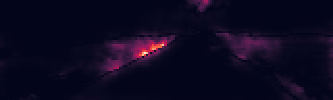}\\
\rowname{0.1}&
\includegraphics[width=.090\linewidth]{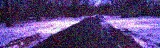}&
\includegraphics[width=.090\linewidth, decodearray={0.25 1 0.25 1 0.25 1}]{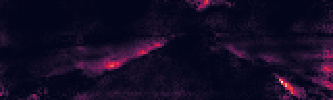}&
\includegraphics[width=.090\linewidth, decodearray={0.25 1 0.25 1 0.25 1}]{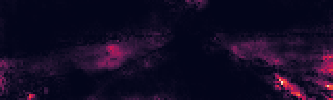}&
\includegraphics[width=.090\linewidth, decodearray={0.25 1 0.25 1 0.25 1}]{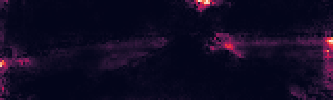}&
\includegraphics[width=.090\linewidth, decodearray={0.25 1 0.25 1 0.25 1}]{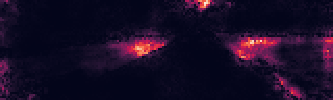}&
\includegraphics[width=.090\linewidth, decodearray={0.25 1 0.25 1 0.25 1}]{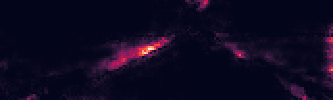}&
\includegraphics[width=.090\linewidth, decodearray={0.25 1 0.25 1 0.25 1}]{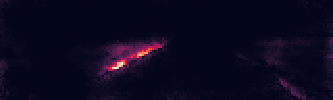}&
\includegraphics[width=.090\linewidth, decodearray={0.25 1 0.25 1 0.25 1}]{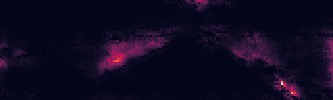}&
\includegraphics[width=.090\linewidth, decodearray={0.25 1 0.25 1 0.25 1}]{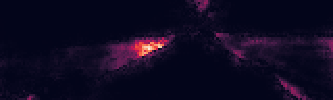}&
\includegraphics[width=.090\linewidth, decodearray={0.25 1 0.25 1 0.25 1}]{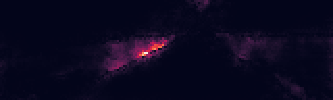}\\
\rowname{0.2}&
\includegraphics[width=.090\linewidth]{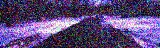}&
\includegraphics[width=.090\linewidth, decodearray={0.25 1 0.25 1 0.25 1}]{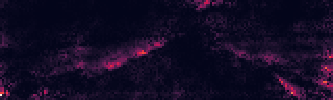}&
\includegraphics[width=.090\linewidth, decodearray={0.25 1 0.25 1 0.25 1}]{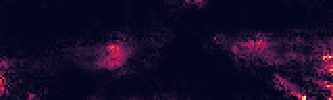}&
\includegraphics[width=.090\linewidth, decodearray={0.25 1 0.25 1 0.25 1}]{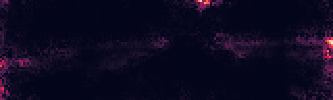}&
\includegraphics[width=.090\linewidth, decodearray={0.25 1 0.25 1 0.25 1}]{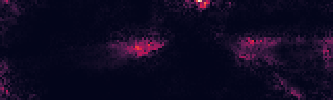}&
\includegraphics[width=.090\linewidth, decodearray={0.25 1 0.25 1 0.25 1}]{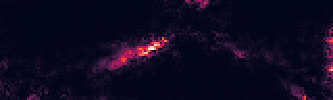}&
\includegraphics[width=.090\linewidth, decodearray={0.25 1 0.25 1 0.25 1}]{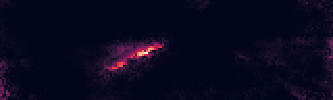}&
\includegraphics[width=.090\linewidth, decodearray={0.25 1 0.25 1 0.25 1}]{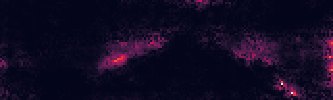}&
\includegraphics[width=.090\linewidth, decodearray={0.25 1 0.25 1 0.25 1}]{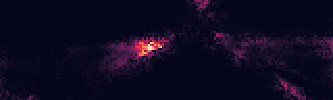}&
\includegraphics[width=.090\linewidth, decodearray={0.25 1 0.25 1 0.25 1}]{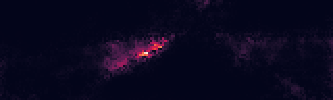}\\
\end{tabular}
\caption{Attention maps in the winter season. In general, the focus shifts from the road to the side of the road compared to summer. Unimportant regions on the off-road are attended by LSTM, GRU and MGU.  A winter input image and its saliency maps of the analyzed models are displayed, with increasing noise of Gaussian noise of $\sigma^2=0.1$ and $\sigma^2=0.2$ variances, in the second and third rows. }%
\label{fig:saliency_winter}
\end{figure}

Figure \ref{fig:saliency_summer} shows our results in the summer season, where lighter-highlighted regions indicate the attention of the network. We found that the LTSM takes into account irrelevant parts of the image, during its decision-making. Similarly, the LC-based models highlight the sides of the picture, too. The rest of the models have most of the attention on the road, and GCU networks especially, focus on the horizon. In the case of winter, which can be seen in Figure~\ref{fig:saliency_winter}, irrelevant features of the images are highlighted by the traditionally used gated RNNs, and LC-SA shows also less focused attention to the contour of the road compared to LRCs.

By injecting extra Gaussian noise which was not present during training into the tests, we can measure how robust their attention is, that is, how much attention changes compared to the natural noise-free setting. The noisy attention maps are displayed in the second and third rows of Figures~\ref{fig:saliency_summer}-\ref{fig:saliency_winter}, with zero mean, 0.1 and 0.2 variances, respectively. 

\subsection{Neural Activities}
\begin{figure}
\vspace*{-4mm}
\settoheight{\tempdima}{\includegraphics[width=.2\linewidth]{example-image-a}}%
\centering\begin{tabular}{@{}c@{ }c@{ }c@{ }c@{}c@{}}
&Cell \#1 & Cell \#1 & Cell \#2& Cell \#2 \\
&in summer & in winter & in summer& in winter \\
\rowname{LSTM}&
\includegraphics[width=.13\linewidth]{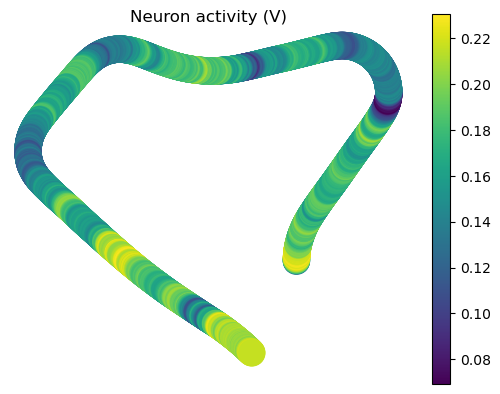}&
\includegraphics[width=.13\linewidth]{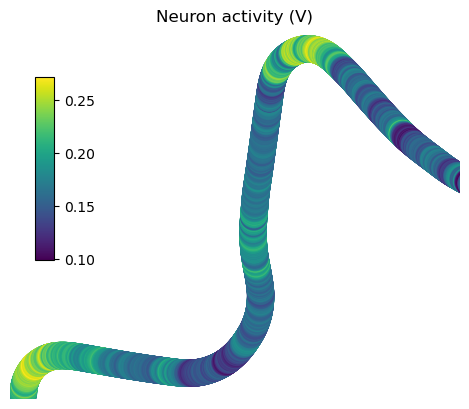}&
\includegraphics[width=.13\linewidth]{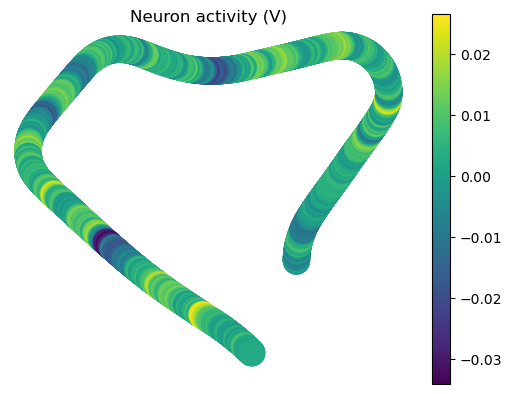}&
\includegraphics[width=.13\linewidth]{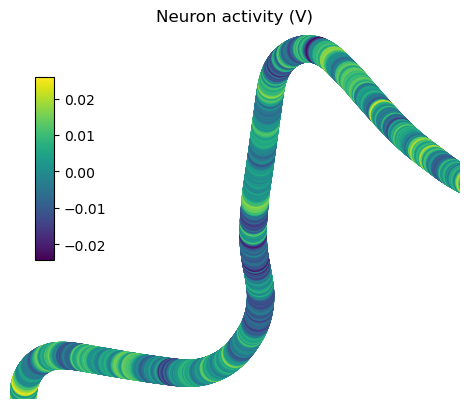}\\
\rowname{GRU}&
\includegraphics[width=.15\linewidth]{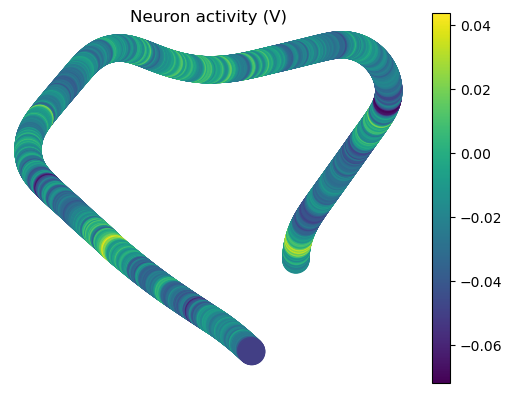}&
\includegraphics[width=.13\linewidth]{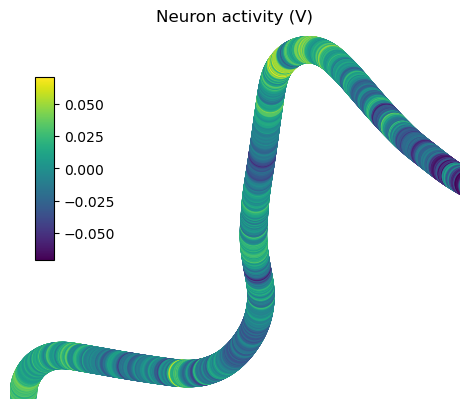}&
\includegraphics[width=.13\linewidth]{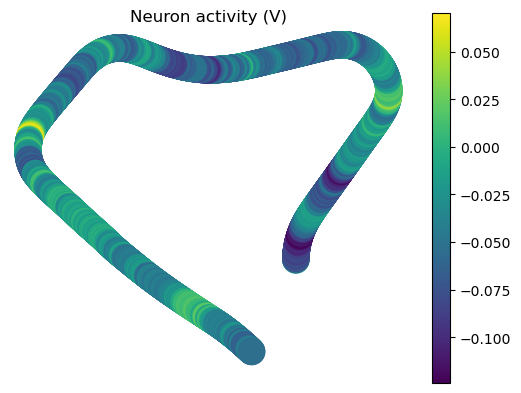}&
\includegraphics[width=.13\linewidth]{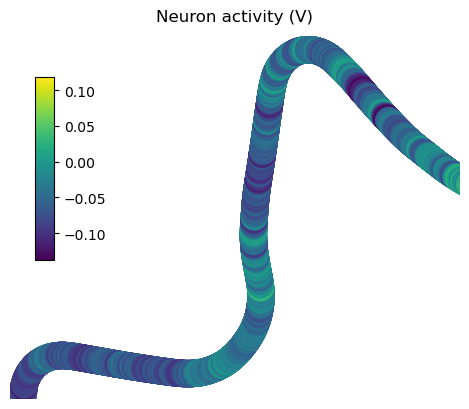}\\
\rowname{MGU}&
\includegraphics[width=.13\linewidth]{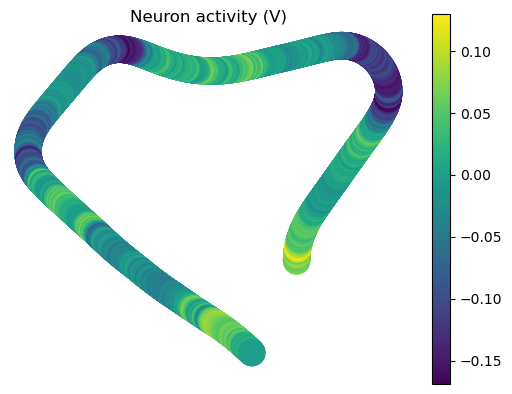}&
\includegraphics[width=.13\linewidth]{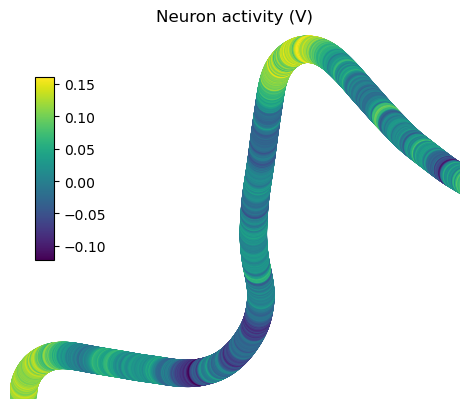}&
\includegraphics[width=.13\linewidth]{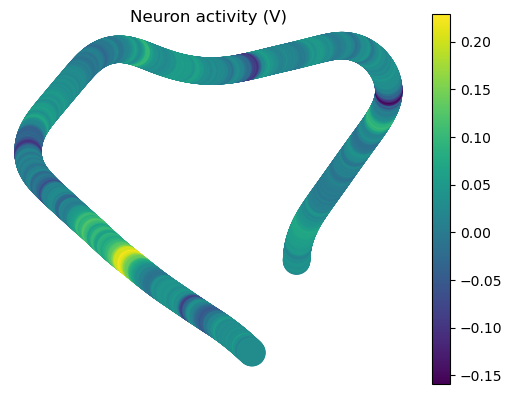}&
\includegraphics[width=.13\linewidth]{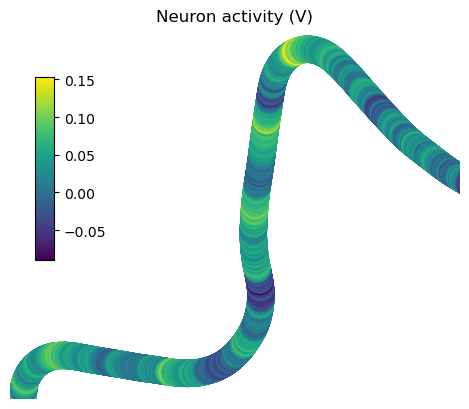}\\
\rowname{CT-RNN}&
\includegraphics[width=.13\linewidth]{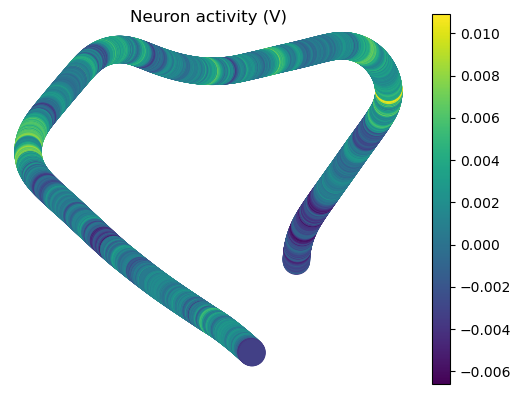}&
\includegraphics[width=.13\linewidth]{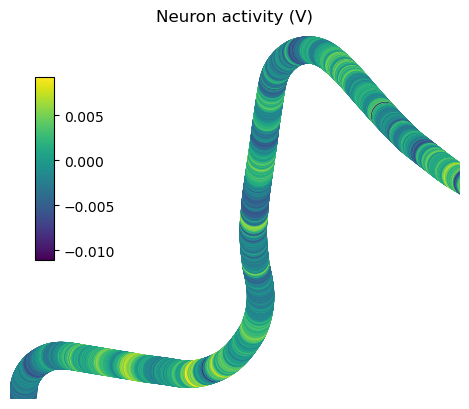}&
\includegraphics[width=.13\linewidth]{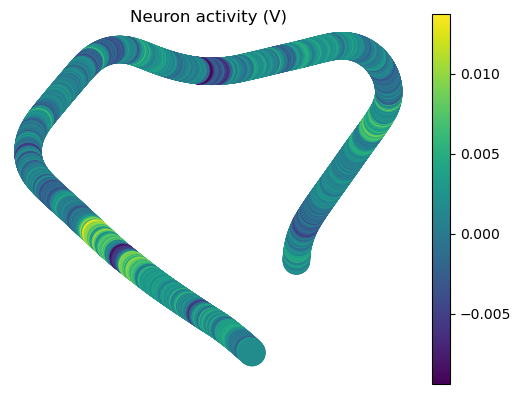}&
\includegraphics[width=.13\linewidth]{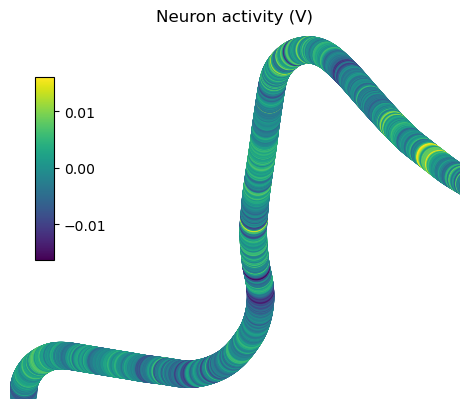}\\
\rowname{LTC}&
\includegraphics[width=.13\linewidth]{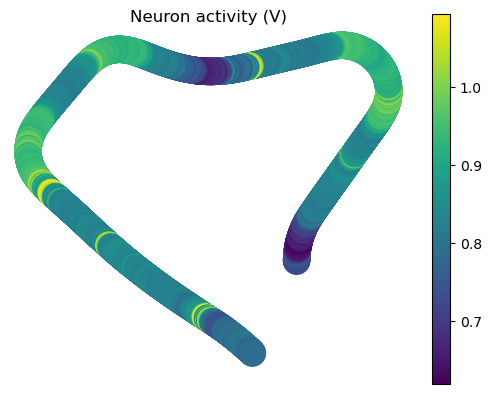}&
\includegraphics[width=.13\linewidth]{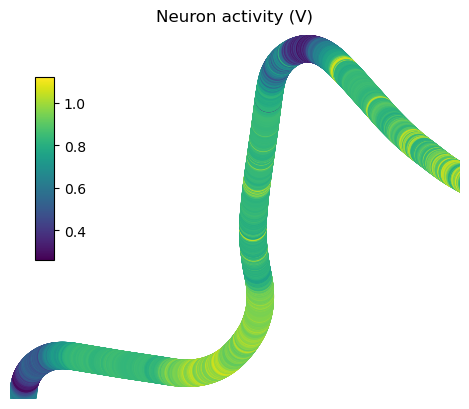}&
\includegraphics[width=.13\linewidth]{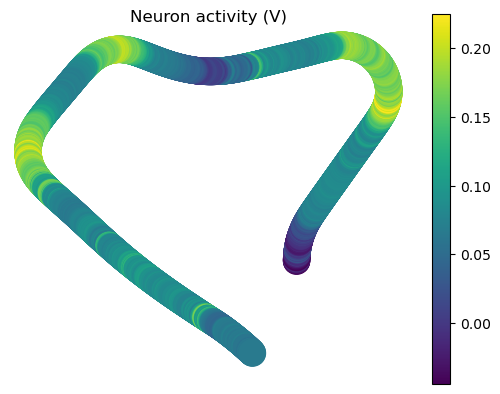}&
\includegraphics[width=.13\linewidth]{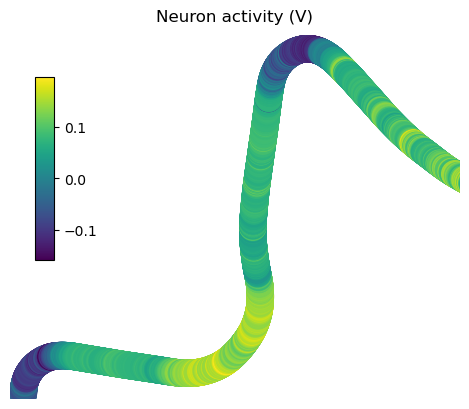}\\
\rowname{LC-NA}&
\includegraphics[width=.13\linewidth]{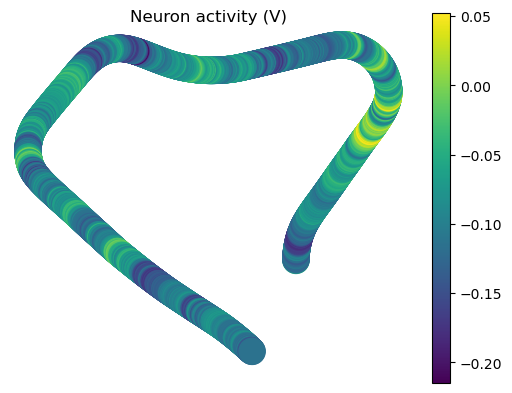}&
\includegraphics[width=.13\linewidth]{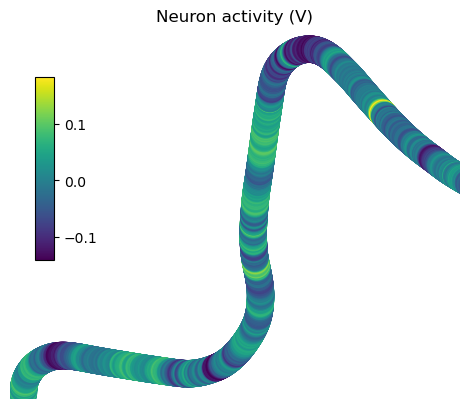}&
\includegraphics[width=.13\linewidth]{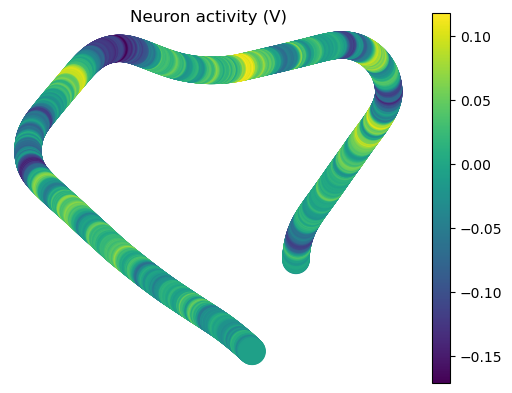}&
\includegraphics[width=.13\linewidth]{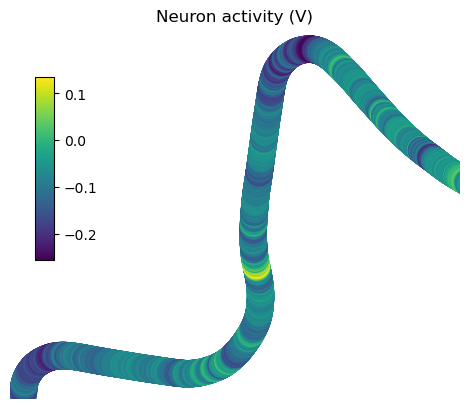}\\
\rowname{LC-SA}&
\includegraphics[width=.13\linewidth]{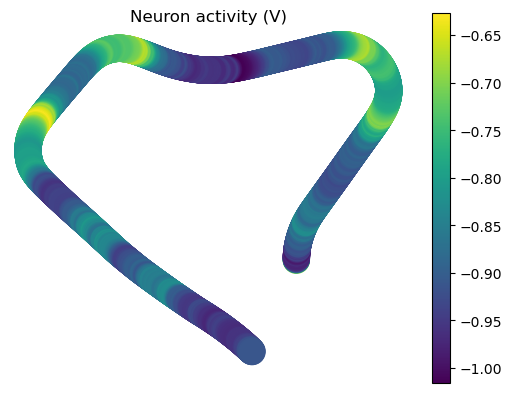}&
\includegraphics[width=.13\linewidth]{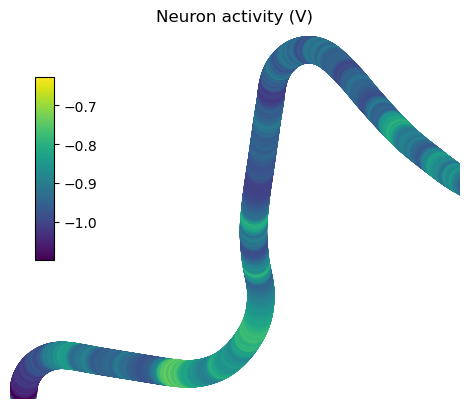}&
\includegraphics[width=.13\linewidth]{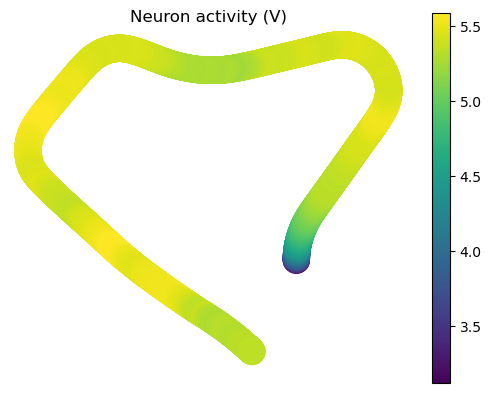}&
\includegraphics[width=.13\linewidth]{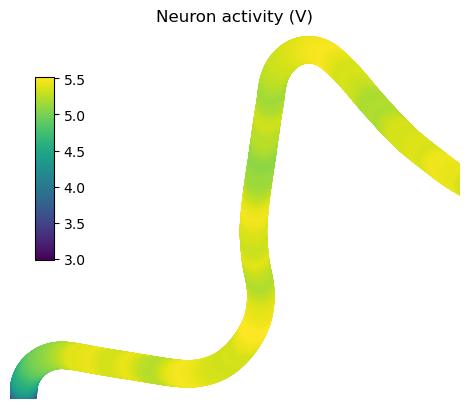}\\
\rowname{LRC-NA}&
\includegraphics[width=.13\linewidth]{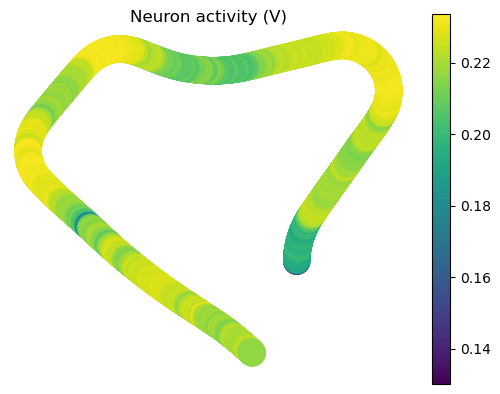}&
\includegraphics[width=.13\linewidth]{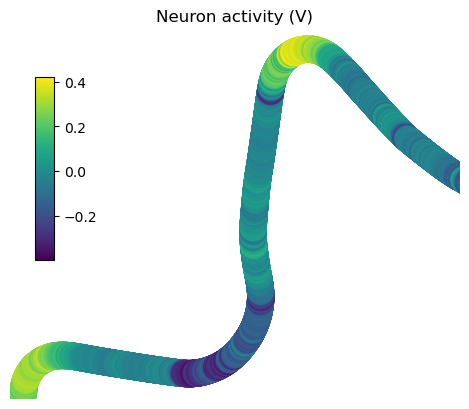}&
\includegraphics[width=.13\linewidth]{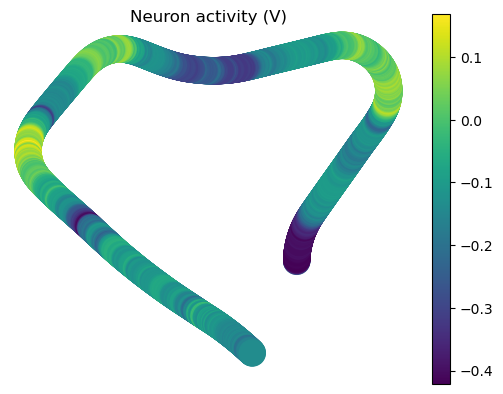}&
\includegraphics[width=.13\linewidth]{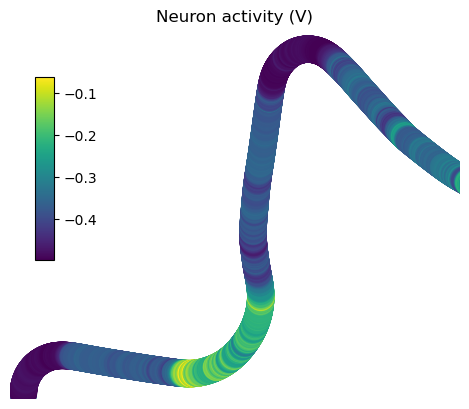}\\
\rowname{LRC-SA}&
\includegraphics[width=.13\linewidth]{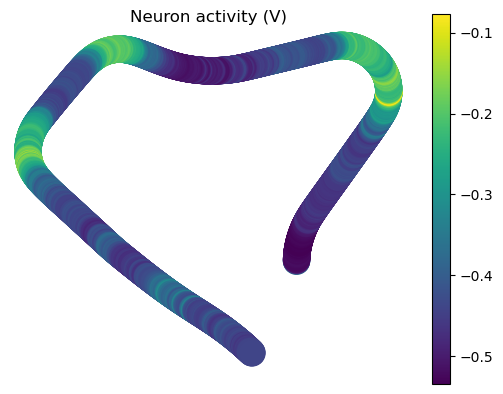}&
\includegraphics[width=.13\linewidth]{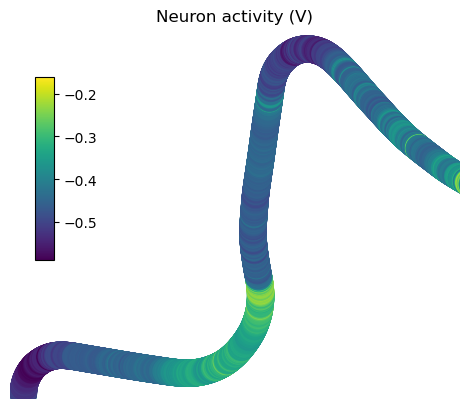}&
\includegraphics[width=.13\linewidth]{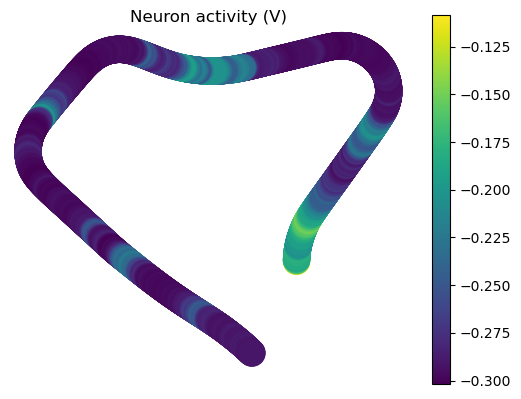}&
\includegraphics[width=.13\linewidth]{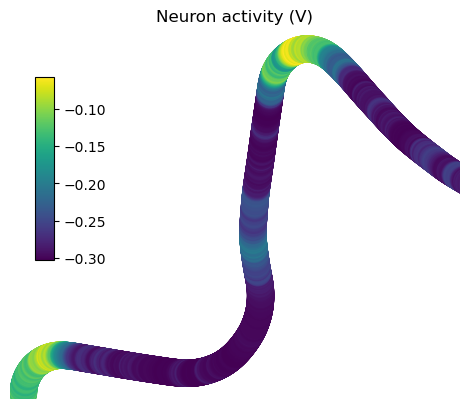}\\
\end{tabular}
\caption{The neural activity of two cells in a Lane-Keeping policy, projected over time on the 1km road driven in both summer and in winter.}
\label{fig:neural_activity}
\end{figure}

Figure~\ref{fig:neural_activity} shows the hidden state activation patterns during closed-loop driving, for three popular gated recurrent units (LSTMs, GRUs, and MGUs) and six biological units (CT-RNN, LTC, LC-NA, LC-SA, LRC-NA, and LRC-SA). The liquid capacitance units considered have either electrical (only LC) or chemical (with liquid resistance, thus LRC) synapses, and neural (NA) or synaptic (SA) activation, respectively. 
While all RNNs can exhibit a neural activity that roughly matches the traversed road geometry, LTC and LRCs display smoother and more identifiable turning patterns during driving, compared to the other models. Finally, LRC-SAs have the most interpretable activity patterns among all models, supported by Table~\ref{tab:correlations}.

\end{document}